\documentclass[sigconf]{acmart}

\setcopyright{acmlicensed}
\copyrightyear{2026}
\acmYear{2026}
\acmDOI{XXXXXXX.XXXXXXX}
\acmConference[RecSys '26]{Make sure to enter the correct
  conference title from your rights confirmation email}{June 03--05,
  2018}{Woodstock, NY}
\acmISBN{978-1-4503-XXXX-X/2018/06}

\usepackage{xspace}
\usepackage{enumitem}
\usepackage{multirow}
\usepackage{colortbl}
\usepackage{nicefrac}

\usepackage{graphicx}
\usepackage[T1]{fontenc}
\usepackage{textcomp}
\usepackage{hyperref}
\usepackage{pifont}
\usepackage{makecell}
 
\usepackage[center]{subfigure}
\usepackage{enumitem}
\usepackage{amsfonts}
\newcolumntype{s}{>{\columncolor{gray!10}} c}

\begin{document}

\title{Understanding ID-Text Complementarity in Sequential Recommendation}

\author{Liam Collins, Bhuvesh Kumar, Clark Mingxuan Ju, Tong Zhao, Donald Loveland, Leonardo Neves, Neil Shah}
\email{{lcollins2,bkumar,mju,tzhao,dloveland,lneves,nshah}@snapchat.com}
\affiliation{%
  \institution{Snap Inc.}
  \city{Santa Monica}
  \state{CA}
  \country{USA}
}

\renewcommand{\shortauthors}{Liam Collins et al.}
\renewcommand{\shorttitle}{ID and Text Complementarity in SR}

\begin{abstract}
Dense retrieval-based Sequential Recommendation (SR) systems increasingly leverage text features to represent items, and generally do so in one of two ways:  (i) by completely replacing ID embeddings with text embeddings, or (ii) by carefully combining ID and text embeddings through complex fusion mechanisms. While these lines of work have produced strong recommendation performances, they offer conflicting perspectives on ID-text {\em complementarity}, or the extent to which ID and text embeddings specialize in modeling different signals: the former suggests a {\em lack} of complementarity, and the latter argues it {\em exists but must be harnessed carefully}. Moreover, neither workstream conducts an in-depth study of ID-text complementarity. We aim to clarify this picture by developing a rigorous understanding of the complementarity of ID- and text-based SR models in this work. Our study reveals that these models {\em do} learn complementary signals, meaning that either should provide performance gain when used properly alongside the other. Motivated by this, we introduce and evaluate a new, simple SR baseline that preserves ID-text complementarity through independent model training, then harnesses it via ensembling. Despite this method's simplicity, we show it outperforms several competitive SR baselines, implying a third perspective on ID-text complementarity: both features are necessary to achieve state-of-the-art SR performance, but complex fusion strategies are not.
\end{abstract}






\maketitle

 
\section{Introduction} 

Sequential Recommendation (SR) is an increasingly popular paradigm within Recommendation Systems (RecSys) that aims to predict the next item(s) a user will interact with given the user's prior interaction sequence \cite{hidasi2015session,tang2018personalized,kang2018self,sun2019bert4rec,rajput2023recommender}. 
Since most SR approaches model users purely through their interaction histories, the method for representing items in those histories is of utmost importance to model performance. 
Foundational SR approaches compute item representation solely from items' IDs, training free-parameter  embedding tables indexed by item ID \cite{hidasi2015session,tang2018personalized,kang2018self,sun2019bert4rec}. More  recently, SR methods have leveraged advancements large language models to  incorporate information from items' text features, such as their title and/or descriptions \cite{hou2022towards,yuan2023go,zhang2024id}.

The emergence of text features raises important questions around how to position these features  relative to ID features in SR pipelines.
Several works that use dense (embedding-based) retrieval to solve SR completely replace ID embeddings with text embeddings,   arguing that under certain conditions text embeddings can match or exceed the performance of ID embeddings \cite{yuan2023go,li2023exploring,sheng2024language}.
On the other hand, a number of works show that ID embeddings {can} provide value when used in addition to text features, but these methods often rely on complicated ID-text fusion strategies such as careful alignment architectures \cite{liu2024llm,ren2024representation,liu2024alignrec} and/or multi-stage training  \cite{li2024enhancing,rajput2023recommender,yang2024unifying}.
Taken together, these two lines of work cast doubt on the possibility of easily extracting value from both ID and text features in the same model.

Underpinning this doubt is a lack of thorough understanding of the {\em complementarity} of ID and text-based SR models, i.e. the \textit{extent to which they learn different signals}. Understanding complementarity is critical to gauge the marginal value each feature type can provide when used in concert with the other, and can inform efficient ways of doing so. Yet, to our knowledge the literature has not rigorously studied ID-text complementarity besides in showing that text-based models generalize better to cold-start items \cite{yuan2023go,sheng2024language,li2023exploring}.

To address this gap, we empirically  analyze ID- and text-based SR model complementarity, focusing on  dense retrieval-style SR for simplicity.
Specifically, we contribute:
\begin{itemize}[leftmargin=*] 

\item \textbf{User-based complementarity metrics.} We introduce metrics quantifying the complementarity of two SR models based on the similarity of the sets of users that each model performs well on. These metrics develop a general framework for understanding model complementarity.

\item \textbf{Fine-grained ID-text complementarity understanding.} We quantify  ID- and text-based SR model complementarity via our proposed metrics. For the most part, we find strong ID-text complementarity relative to baseline model pairs that differ in ways other than their use of IDs or text, suggesting the relative significance of ID-text complementarity.

\item \textbf{Understanding complementarity through ensembling.} 
To further understand ID-text complementarity in terms of its immediate utility, we study the performance of  a simple ensembling baseline termed \texttt{EnsRec}. \texttt{EnsRec} trains ID- and text-based SR models independently, then  \texttt{Ens}embles their predictions at inference. Surprisingly, we show empirically that this simple approach outperforms a multitude of competitive methods, affirming the strength of ID-text complementarity and calling into question the need for complicated methods to exploit it.
\end{itemize}

\noindent Our code is available at: 
\url{https://github.com/snap-research/EnsRec}.

\section{Related Work}

\noindent \textbf{Item Embeddings in SR.} 
Like in traditional RecSys, Sequential Recommendation (SR) has  historically relied solely on ID embeddings to represent items \cite{hidasi2015session,tang2018personalized,kang2018self,sun2019bert4rec}.
However, in recent years, advancements in language and vision modeling have motivated  a plethora of methods for incorporating modality (i.e. textual or visual) features in item representations, including directly replacing ID embeddings with modality embeddings \cite{yuan2023go,zhang2024id,hou2022towards,li2023exploring}, fusing ID and modality embeddings \cite{hu2025alphafuse,liu2024llm,liu2024alignrec,li2024enhancing}, computing semantic IDs from modality embeddings \cite{rajput2023recommender,yang2024unifying}, and using item text descriptions to directly query an LLM for recommendations \cite{bao2023tallrec,li2023text}.
Here, for a controlled complementarity analysis, we utilize text embeddings by simply replacing ID embeddings in the SR model with projected text embeddings \cite{yuan2023go}.
Besides proposing new methods, several works \cite{yuan2023go,sheng2024language,li2023exploring}
compare the performances of ID and text embeddings  and highlight the superiority of text embeddings for cold-start scenarios, but do not otherwise study their complementarity.
In fact, \cite{sheng2024language} claims that ID and text embeddings can be linearly related, suggesting a {\em lack} of complementarity.
To our knowledge, we are the first to quantify the complementarity of ID- and text-based SR models along axes other than cold-start performance.

 \vspace{-1mm}

\section{Methodology} \label{sec:method}

\subsubsection{Preliminaries} \label{sec:prelim}
Let $\mathcal{I}$ be the set of items and $\mathcal{U}$ be the set of users. 
Each user $u \in \mathcal{U}$ has an interacted item sequence $ \mathcal{S}_{:n}^{(u)} := (i_{1}^{(u)}, \dots, i_{n}^{(u)}) \in \mathcal{I}^{\otimes n}$, where for simplicity, we assume in this section that $n$ is the full sequence length for all users, and the $n$-th item is the test target. 
We consider dense retrieval SR models of the form  $M := T \circ E$, which for any $k\in \{1, \dots, n-1\}$ map an item sequence $\mathcal{S}_{:k}^{(u)}$  first to an item embedding sequence $E(\mathcal{S}_{:k}^{(u)}):=(E(i_{1}^{(u)}), \dots, E(i_{k}^{(u)})) \in (\mathbb{R}^d)^{\otimes k}$, then  to a single embedding $ T(E(\mathcal{S}_{:k}^{(u)})) = M(\mathcal{S}_{:k}^{(u)})\in \mathbb{R}^d$. Here, $E$ is an ID-indexed embedding table, or a language model that acts on item text features, and $T$ is a sequence model, e.g. a transformer. Finally, the model outputs a 
score function $s_{M, u, k}: \mathcal{I} \rightarrow \mathbb{R}$, where by default, $s_{M, u, k}(i) = \text{sim}(M(\mathcal{S}_{:k}^{(u)}), E(i))$ and $\text{sim}()$ is cosine similarity. For any such score function $s$ and item $i \in \mathcal{I}$, let $\text{rank}_{\mathcal{I}}(s, i)$ be the rank of $s(i)$ among all item scores, in descending order.
Further, define $$\mathcal{C}_{M} := \{ u \in \mathcal{U}: \text{rank}_{\mathcal{I}}(s_{M, u, n-1}, i^{(u)}_{n}) \leq 10\} $$ as the set of users for which $M$ successfully achieves test Recall@10. 




\subsection{Quantifying Complementarity} \label{sec:predsim}



 We   study the  complementarity of a pair of  SR models $M$ and $M'$ by comparing the sets of users that they model well, i.e. $\mathcal{C}_M$ and $\mathcal{C}_{M'}$, respectively. The more dissimilar these sets are, the more complementary $M$ and $M'$ are.
We measure these sets' similarity using the Jaccard Index, namely:
\begin{align}
  \tilde{J}({M}, {M'}) :=   J(\mathcal{C}_{M}, \mathcal{C}_{M'}) := \frac{|\mathcal{C}_{M}\cap\mathcal{C}_{M'}|}{|\mathcal{C}_{M}\cup\mathcal{C}_{M'}|},  
\end{align}
i.e. the ratio of the number of of users that {\em both} models perform well on to the number of users that {\em either} model performs well on. The smaller this ratio, the more complementary the models.

We also study $\mathtt{Genie}({M}, {M'}) := |\mathcal{C}_{M}\cup\mathcal{C}_{M'}|/|\mathcal{U}|$, which is the Recall@10 of a "Genie" that for each user knows the test target item and chooses to use the model ($M$ or $M'$) that ranks this item higher. This metric estimates the maximum possible performance of combining two models, since it well-models {\em all} users that {\em either} $M$ or $M'$ well-models. Model pairs with higher complementarity have higher $\mathtt{Genie}$ scores relative to their individual Recall@10's.

\begin{figure}
    \centering
        \includegraphics[width=0.99\linewidth]{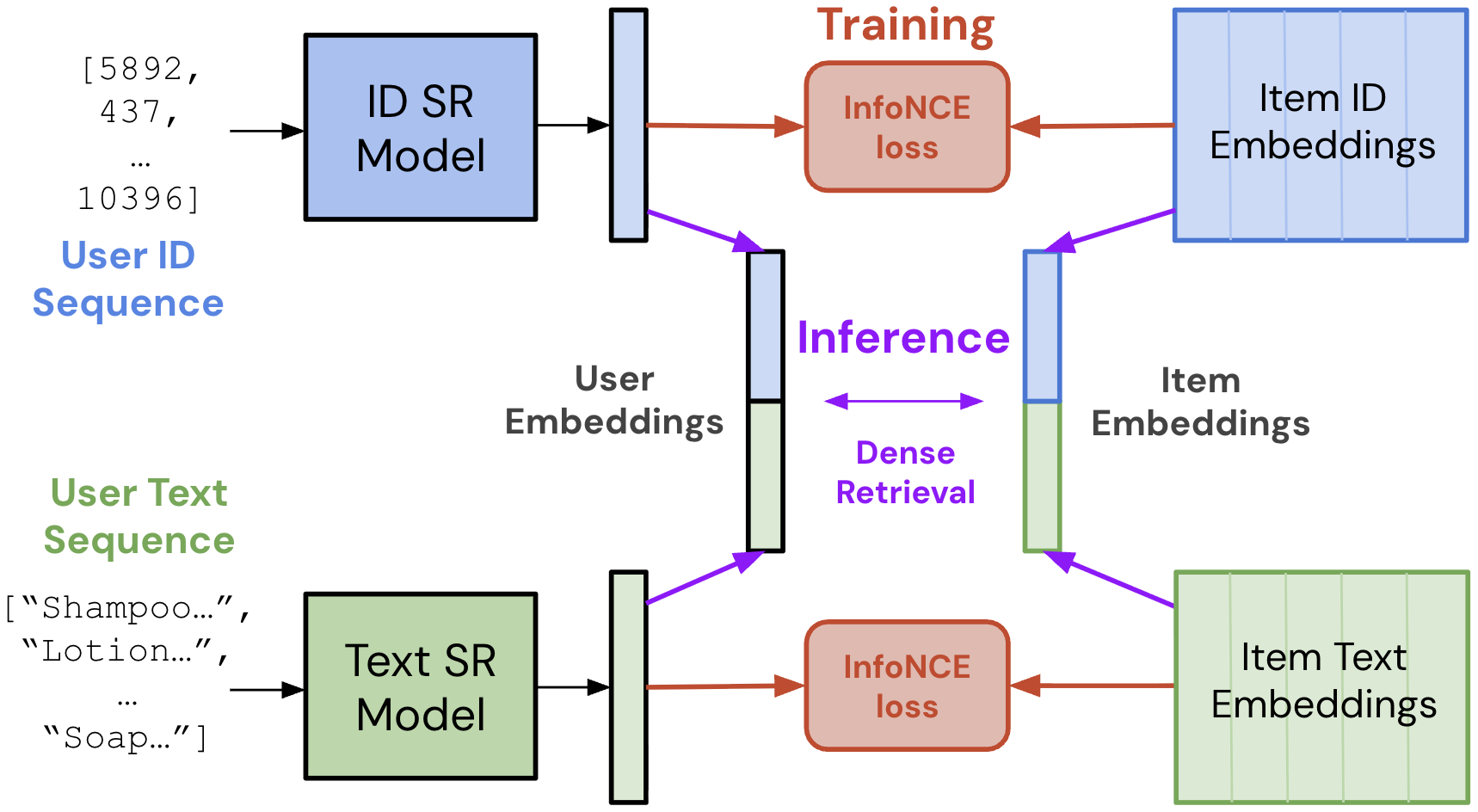}
    \caption{
$\mathtt{EnsRec}$ design. $\mathtt{EnsRec}$ trains ID and Text SR models independently, the performs dense retrieval on the concatenated normalized embeddings at inference.
    }
\label{fig:model}
\end{figure}

\subsection{EnsRec: A Simple Baseline Model}
While $\mathtt{Genie}$ is an analysis tool that  {\em upper bounds} the performance of a model that optimally leverages complementarity, we next propose an implementable method that {\em lower bounds} this performance.
We introduce \texttt{EnsRec}, an SR method that  exploits ID-text complementarity through simple \texttt{Ens}embling.
Let $M_{\text{ID}} := T \circ E_{\text{ID}}$ and $M_{\text{text}}:= T' \circ E_{\text{text}}$ denote ID- and text-only SR models that embed items with an ID-indexed embedding table ($E_{\text{ID}}$) and a frozen language model ($E_{\text{text}}$), respectively. Here $T$ and $T'$ are instances of the same transformer architecture with independent weights. $\mathtt{EnsRec}$ independently trains $M_{\text{ID}}$ and $M_{\text{text}}$, then ensembles their predictions during inference as:
\begin{align}
s_{\tilde{M}_{\text{Ens}}, u, k}(i)  = s_{M_{\text{ID}}, u, k}(i) + s_{M_{\text{text}}, u, k}(i),
\end{align}
where we use $\tilde{M}_{\text{Ens}}$ to denote the $\mathtt{EnsRec}$ model, and $s_{M_{\text{text}}, u, k}$ and $s_{M_{\text{text}}, u, k}$ are score functions as defined in Section \ref{sec:prelim}.

\noindent \textbf{Leveraging complementarity.} $\mathtt{EnsRec}$ first preserves the complementarity of $M_{\text{ID}}$ and $M_{\text{text}}$ by training them independently. Then, it aims to exploit this complementarity by recalling items that either model is very confident in, and {\em not} recall items that either model is very doubtful of. It tries to achieve this via score summation, prioritizing simplicity and efficiency over performance. 

\noindent \textbf{Implementation.} In large-scale settings, we can compute the top-ranked items over $s_{\tilde{M}_{\text{Ens}}, u, k}(\cdot)$ by normalizing and concatenating the user embeddings $M_{\text{ID}}(\mathcal{S}^{(u)}_{:k})$ and $M_{\text{text}}(\mathcal{S}^{(u)}_{:k})$, and likewise for the item embeddings $ E_{\text{ID}}(i)$ and  $E_{\text{text}}(i)$ for all $i\in \mathcal{I}$. Then, for each user, we run efficient ANN search over item embeddings to recommend items\footnote{This makes $\mathtt{EnsRec}$ equivalent, up to embedding normalization, at inference time to the popular SR method FDSA \cite{zhang2019feature}. The only difference during training is that FDSA jointly trains the ID and text embeddings. Note that $\mathtt{EnsRec}$ and FDSA have the same computational cost.}.

\section{Experiments}

We empirically study the following research questions:
\begin{itemize}[leftmargin=*]
    \item \textbf{Q1.} {\em How complementary} are ID- and text-based SR models?
    \item \textbf{Q2.} How {\em immediately useful} is this complementarity, as measured by $\mathtt{EnsRec}$'s performance?
    \item \textbf{Q3.} How {\em sensitive} is this performance to the ensembling method?
\end{itemize}


\subsection{Experimental Settings}

\subsubsection{Datasets} We use several next-item prediction benchmarks: Beauty, Toys and Games (Toys), and Sports and Outdoors (Sports) from the Amazon Review 2018 dataset~\citep{he2016ups}, and the Steam video game dataset  \cite{kang2018self}. 
We use the same preprocessing for Amazon as in several SR works, e.g. \cite{rajput2023recommender,yang2024unifying,ju2025generative,hu2025alphafuse} and for Steam as in \cite{yang2024unifying}, both entailing 5-core filtering and leave-one-out splitting. 
The  Amazon text features are title, price, categories, and description, and title, genre, specs, tags, price, and publisher for Steam.
 We use frozen SentenceT5-XXL \cite{ni2021sentence} to compute of these features, as in \cite{yang2024unifying}.

\subsubsection{Baselines}
We consider several ID and/or text-based SR methods to study Q2.
\textbf{ID-Only} and \textbf{Text-Only} are the single-encoder versions of $\mathtt{EnsRec}$,  
and are analogous to SASRec \cite{kang2018self} and {MoRec} \cite{yuan2023go}, respectively.  \textbf{LLMInit} \cite{harte2023leveraging,hu2024enhancing,hu2025alphafuse} is ID-Only with initialization via a singular value decomposition of the text embeddings.
\textbf{WhitenRec} \cite{zhang2024id} and \textbf{UniSRec} \cite{hou2022towards} use only text embeddings with fixed and trainable whitening layers, respectively.
\textbf{RLLMRec-Con} and \textbf{RLLMRec-Gen} \cite{ren2024representation} distill textual information into ID embeddings.
\textbf{LLM-ESR} \cite{liu2024llm} employs ID-text cross attention, ID and text sequence encoders, an item alignment loss, and user self-distillation.
\textbf{AlphaFuse} \cite{hu2025alphafuse} trains  ID embeddings in the text embedding  null space.
\textbf{LIGER-Dense} \cite{yang2024unifying} adds ID and text embeddings prior to sequential encoding, and
\textbf{LIGER} \cite{yang2024unifying} and \textbf{TIGER} \cite{rajput2023recommender} employ semantic IDs and generative retrieval \cite{ju2025generative}.
\textbf{FDSA} \cite{zhang2019feature} is most similar to $\mathtt{EnsRec}$, differing only in jointly training the concatenated ID and text user and item embeddings instead of  each independently.

\subsubsection{Implementation Details}
We use the implementations of TIGER, LIGER and LIGER-Dense  from \cite{yang2024unifying}, and implement all other methods using the same base model architecture and core training loss as in \cite{yang2024unifying} (and \cite{rajput2023recommender}), borrowing some code from \cite{hu2025alphafuse} to do so. 
All sequence encoders are T5 encoder-only models \cite{raffel2020exploring} with 6 layers, 6 heads per layer, feed-forward dimension 1024 and key-value dimension 64. $\mathtt{EnsRec}$ and FDSA use ID and text embedding dimensions 128, meaning their effective user and item embedding dimensions are 256 after concatenation. AlphaFuse uses 256-dimensional text embeddings and adds $128$-dimensional ID embeddings to their lowest-energy subspace.
We tune the ID and/or text embedding dimension for all other methods in $\{128,256\}$, thus $\mathtt{EnsRec}$ does not unfairly benefit from larger embedding dimension. AlphaFuse, WhitenRec, and LLMInit obtain their 256-dimensional text embeddings through a singular value decomposition (SVD) of the original 768-dimensional embeddings, while the other approaches train a  projection module.

We additionally tune three hyperparameters for all methods: learning rate in $\{0.0001, 0.0003, 0.001 \}$, embedding dropout rate in $\{0.1, 0.5\}$, and, when applicable, the text projection module in $\{\text{linear}, \text{3-layer MLP}\}$, and otherwise
adopt the hyperparameters from \cite{yang2024unifying}.
 All approaches use the full-batch InfoNCE loss \cite{oord2018representation} for next-item prediction, 
and reported  weights for any auxiliary losses. We early stop on validation Recall@10.
We run all experiments on one 40GB NVIDIA A100 GPU.

\subsection{Q1: How complementary?} \label{sec:rq1}
We start by quantifying ID- and text-based SR model complementarity using the user-based metrics defined in Section  \ref{sec:predsim}. 
To gauge this complementarity's significance, we also measure the complementarity of pairs of SR models that differ in ways other than the  feature type they use. 
In particular, we consider 
$M_{\text{ID}}$ and  $M_{\text{text}}$ as the base ID- and Text-only models, respectively. $M_{\text{ID-$T$}}$ and $M_{\text{text-$T$}}$  are variants of these that  change the sequence encoder $T$ from the base T5 encoder to a 2-layer, 1-head/layer decoder, as in SASRec \cite{kang2018self}. Similarly, $M_{\text{ID-$\ell$}}$ and $M_{\text{text-$\ell$}}$  use in-batch negative samples instead of full-batch in the InfoNCE loss, $M_{\text{ID-init}}$ initializes the ID embeddings using a random initialization with 25$\times$ the base variance, and $M_{\text{text-$E$}}$ replaces SentenceT5-XXL embeddings with Sup-SimCSE-BERT \cite{gao2021simcse} embeddings.

Table \ref{tab:rq11} compares the complementarity of pairs of these models, with all results averaged over 3 trials.
Besides pairs that involve models trained with the in-batch loss,  the {\em (ID, text) pair is  more complementary than (ID, ID-$^*$) and (text, text-$^*$)  pairs} according to the Jaccard Index $\tilde{J}$ of the models' sets of correct users.  The {\em (ID, text) pair also achieves the best $\mathtt{Genie}$ scores}, despite some of the other pairs having larger sum of individual model Recall@10's.

\begin{table}
\caption{ Complementarity metrics. We denote the models $(M_a, M_b)$ as $(a,b)$ for all $(a,b)$. R@10 is the  Recall@10's of $(M_a, M_b)$, and $\mathtt{Genie}$ and $\tilde{J}$ are  $\mathtt{Genie}(M_a, M_b)$ and $\tilde{J}(M_a, M_b)$, respectively. All results are averaged over 3 trials.
}
    \label{tab:rq11}
    \setlength\tabcolsep{3.2pt}
    \scalebox{1}{
    \begin{tabular}{lcccccc} 
        \toprule
     \multirow{2}{*}{$(a, b)$ }  &  \multicolumn{3}{c}{\textbf{Beauty} }  & \multicolumn{3}{c}{\textbf{Steam}}   \\
     \cmidrule(lr){2-4} \cmidrule(lr){5-7}   
       & R@10   
       & $\mathtt{Genie} $  &  $ \tilde{J}  $ &   R@10
       & $\mathtt{Genie}$   &  $\tilde{J}  $   \\
        \midrule
        (${\text{ID}}, {\text{ID-$T$}}$) 
        & (8.21,  ${7.65}$)  & 10.2  & 0.57  & (19.8, 19.5)  & 22.1  & 0.78   \\
        (${\text{ID}}, {\text{ID-$\ell$}}$)  & (${8.21},$ ${6.94}$)  & 10.1 & 0.53 & (19.8, 14.9) & 21.2 & \underline{0.64} \\
        (${\text{ID}}, {\text{ID-init}}$)  & (${8.21} $, ${8.31}$)  & 10.2 & 0.64 & (19.8, 19.5) & 22.1 & 0.78 \\
        (${\text{text}}, {\text{text-$T$}}$)  & (${9.37} $, ${9.34}$)  & 12.2 & 0.54 & (20.1, 20.0) & \underline{22.4} & 0.78 \\
        (${\text{text}}, {\text{text-$\ell$}}$)  & (${9.37} $, ${6.78}$)  & 12.1 & \textbf{0.33} & (20.1, 15.4) & 21.9 & \textbf{0.61} \\  
        (${\text{text}}, {\text{text-$E$}}$)  & (${9.37} $, ${9.16}$)  & 12.2 & 0.52 & (20.1, 19.6) & 22.2 & 0.78 \\
        \midrule
        \rowcolor{gray!10}
        (${\text{ID}}, {\text{text}}$) &  (${8.21}$, 9.37) & \textbf{12.2} & \underline{0.47} & (19.8, 20.1) & \textbf{22.7} & 0.75 \\
        \bottomrule
    \end{tabular}}
\end{table}

\begin{table*}[htb]
    \centering
    \caption{Average metrics across three trials. The best result(s) (excluding $\mathtt{Genie}$) are \textbf{bold}-typed and the second best are underlined. $^*$ denotes statistically significant improvements over the second-best result with $p<0.05$. 
    }
    \label{tab:overall}
    \setlength{\tabcolsep}{5.8pt}     
    \begin{tabular}{l *{8}{c}} 
        \toprule
        \multirow{2}{*}{\textbf{Method}} & \multicolumn{2}{c}{\textbf{Beauty}} & \multicolumn{2}{c}{\textbf{Sports}} & \multicolumn{2}{c}{\textbf{Toys}} & \multicolumn{2}{c}{\textbf{Steam}} \\
        \cmidrule(lr){2-3} \cmidrule(lr){4-5} \cmidrule(lr){6-7} \cmidrule(lr){8-9}
        & NDCG@10 & Recall@10 &  NDCG@10 & Recall@10 &  NDCG@10 & Recall@10  & NDCG@10 & Recall@10 \\ 
        \midrule
        ID-Only & 4.893 & 8.211 & 2.785 & 4.858 & 5.460 & 8.812 & 15.70 & 19.79 \\
        Text-Only &  4.790 & 9.373 & 2.927 & 5.667 & 4.860 & 9.853 & 15.73  & 20.07 \\
        WhitenRec  & 4.532 & 9.234 & 2.775 & 5.600 & 4.639 & 9.698 & 15.85 & 20.19 \\
        UniSRec  & 5.033 & 9.674 & 2.885 & 5.536 & 4.913 & 9.889 & 15.91 & 20.19 \\  
        LLMInit & \underline{5.220} & 9.504 & \underline{3.019} & \underline{5.714} & {5.276} & \underline{10.01} & \textbf{16.02} & \underline{20.40} \\
        RLMRec-Con  & 4.953 & 8.230 & 2.744 & 4.823 & 5.552 & 8.948 & 15.76 & 19.86 \\
        RLMRec-Gen  & 4.865 & 8.187 & 2.728 & 4.757 & \underline{5.565} & 9.106 & 15.77 & 19.91 \\
        LLM-ESR & {5.181} & \underline{9.717} & 2.913 & 5.542 & 4.910 & 9.535 &  15.50 & 19.77 \\
        AlphaFuse  & 4.386 & 7.466 & 2.295 & 4.136 & 5.444 & 9.317 & 15.51 & 19.60 \\
        FDSA  & 4.464 & 8.387 & 2.340 & 4.854 & 4.882 & 9.013 & 15.83 & 19.96 \\
        LIGER-Dense & 4.726 & 9.190 & {2.951} & 5.622 & 4.667 & 9.465 & 14.92 & 19.10 \\
        LIGER  & 4.019 & 7.456 & 1.922 & 4.406 & 3.763 & 7.164 & 14.77 & 18.46 \\
        TIGER & 3.205 & 5.988 & 1.981 & 3.806 & 2.934 & 5.753 & 14.53 & 18.35 \\
        \midrule 
        \rowcolor{gray!10}\textbf{EnsRec} & \textbf{5.650}$^*$  & \textbf{9.800}  & \textbf{3.432} & \textbf{6.133}  & \textbf{6.248}$^*$ & \textbf{10.609}$^*$  & \textbf{16.02}  & \textbf{20.45}  \\
        \rowcolor{gray!10}{$\;$ +Improv \%} & +8.2 & +0.9 & +13.7 & +7.3 & +12.3 & +6.0 & 0.0 & +0.2 \\
        \midrule 
       Genie(ID-, Text-only) & 7.097 & 12.20 & 4.480 & 7.833 & 7.599 & 12.87 & 17.34 & 22.73 \\
        \bottomrule
    \end{tabular}
\end{table*}

{
\subsection{Q2: Ensembling performance} \label{sec:rq2}
Table \ref{tab:overall} measures the immediate utility of this complementarity by comparing  $\mathtt{EnsRec}$'s  performance with  the baselines, leading to several observations. First,
$\mathtt{EnsRec}$ significantly improves over ID-Only and Text-Only, further confirming ID-text  complementarity.
More surprisingly, $\mathtt{EnsRec}$ outperforms the other SR baselines in nearly all cases,
evincing that {\em simple ensembling  leverages ID-text complementarity more effectively than prior methods}.

We also observe in Table \ref{tab:overall}  that ID- and Text-Only SR models have complementary strengths in  fine-grained ranking
(NDCG@10) and coarse-grained recall (Recall@10), respectively. This may be because the text embeddings better capture global relationships across items useful for coarse-grained recall, whereas the local structure of the ID embeddings better models precise behavioral relationships. We leave further investigation of this phenomenon to future work.
}

\subsection{Q3: Sensitivity to ensembling method}
We next investigate  $\mathtt{EnsRec}$’s sensitivity to the choice of ensembling method. To do so, we extend $\mathtt{EnsRec}$ to a generic ensembling method $\mathtt{EnsRec}_{\alpha,\tau}$.
Denote $s_{\text{ID}}(i) :=s_{M_{\text{ID}}, u, n-1}(i)$ and $s_{\text{text}}(i) :=s_{M_{\text{text}}, u, n-1}(i)$ as the test scores for user $u$ and item $i$ produced by ID- and Text-Only SR models, respectively. 
$\mathtt{EnsRec}_{\alpha, \tau}$ combines these as\footnote{We also tried normalizing the ID and text scores prior to ensembling but found that this reduced performance in all cases.}:
\begin{align}
    s_{\mathtt{Ens}_{\alpha, \tau}}(i) = \alpha \exp(s_{\text{ID}}(i)/\tau) + (1-\alpha)\exp(s_{\text{text}}(i)/\tau)
\end{align}
for any $\alpha \in [0,1]$ and $\tau\in \mathbb{R}$. In this way, $\alpha$ controls the weight on the ID score vs text score, and $\tau$  interpolates between a max ensembler (small $\tau$) and a sum ensembler (large $\tau$).
Indeed, $\mathtt{EnsRec}$ is equivalent to $\alpha=0.5$ and $\tau \rightarrow \infty$ (since $e^x \stackrel{x\rightarrow 0}{\rightarrow} 1+x$). We approximate this with $(\alpha, \tau) = (0.5, 100)$ in the ablation.

\begin{figure}
    \centering
        \includegraphics[width=0.4955\linewidth]{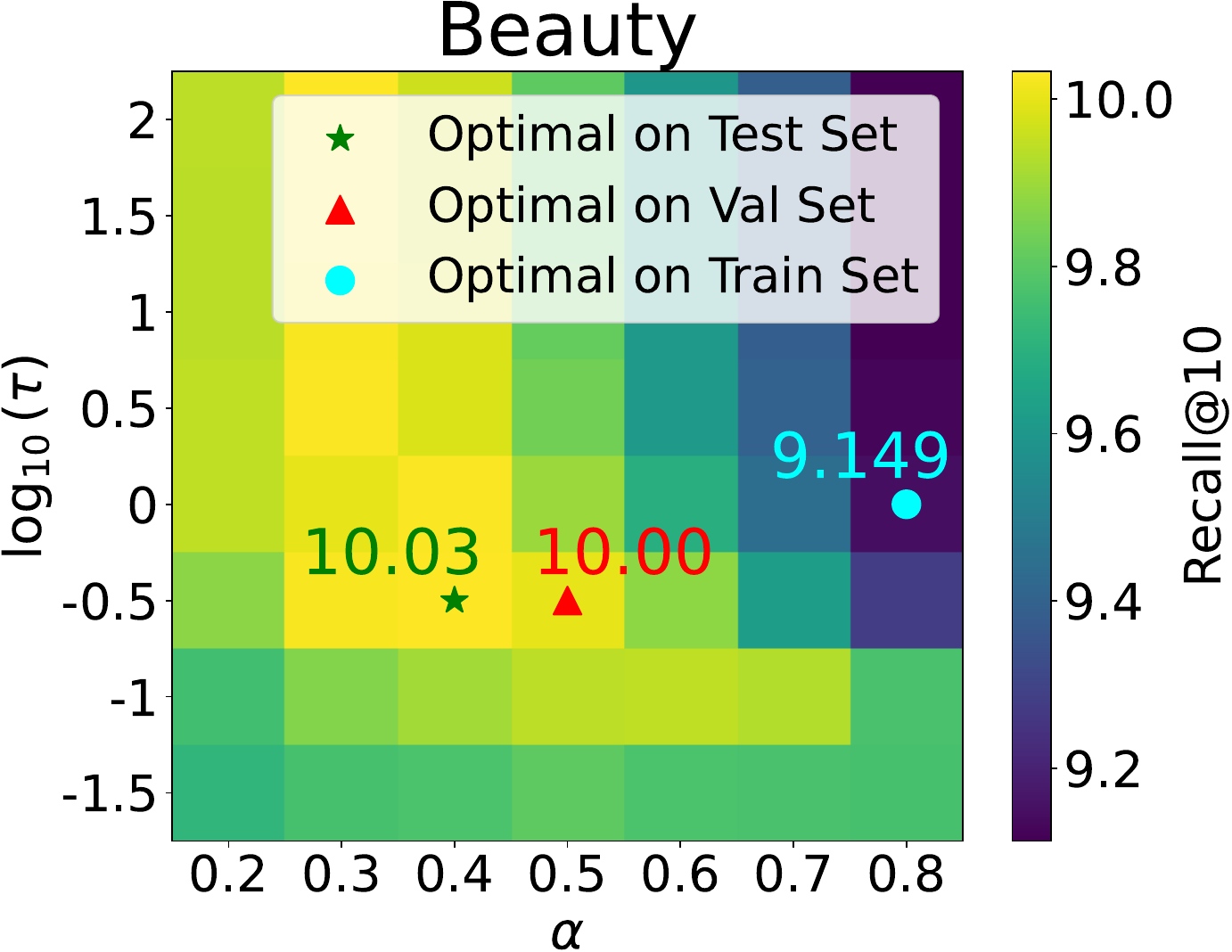}
    \includegraphics[width=0.4955\linewidth]{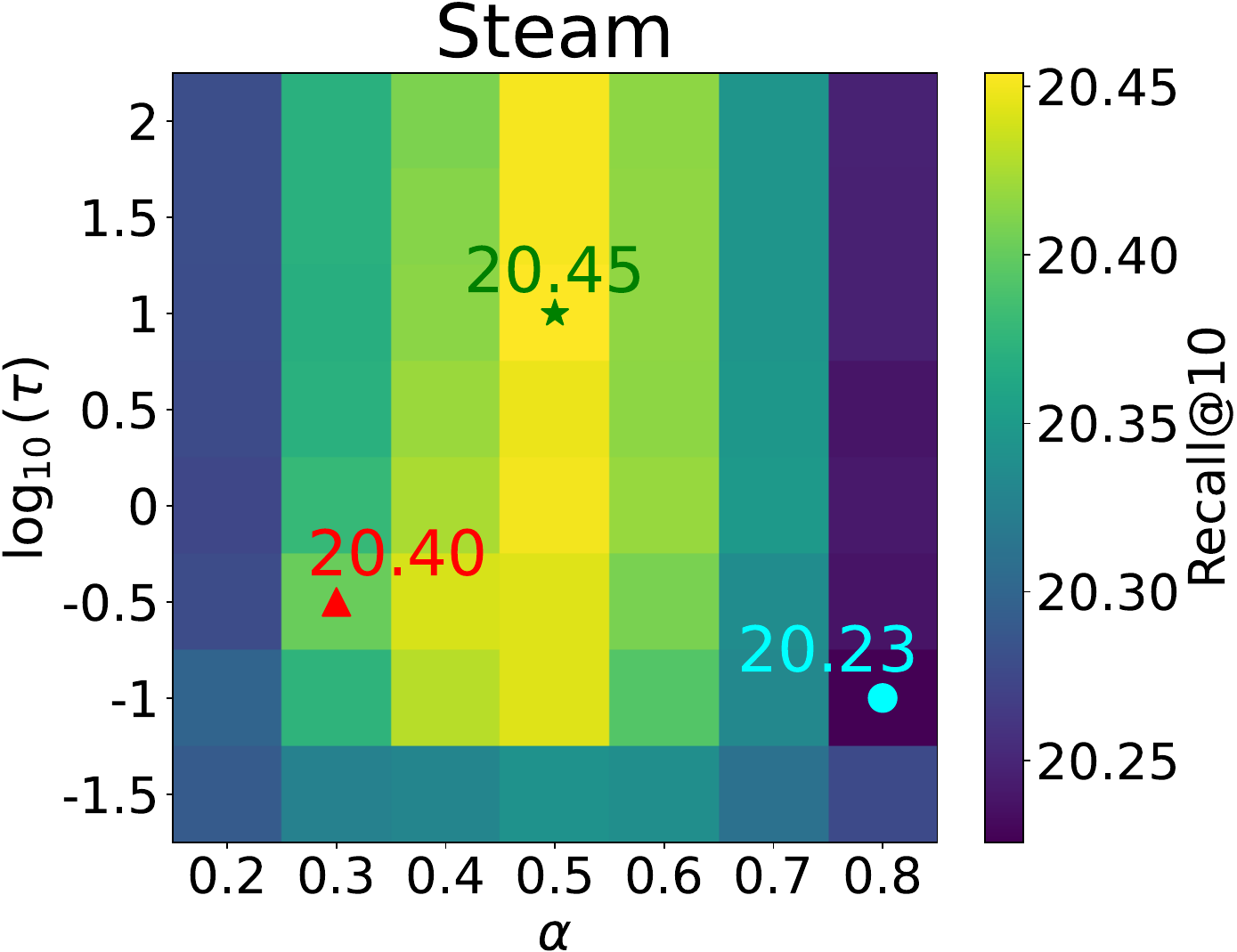}
    \caption{
    Test Recall@10 of $\mathtt{EnsRec}_{\alpha, \tau}$ as a function of $(\alpha, \log_{10}(\tau))$ on (Left) Beauty and (Right) Steam. The $(\alpha, \log_{10}(\tau))$ values that maximize train, validation, and test Recall@10 are plotted along with their corresponding test Recall@10.
    }
\label{fig:1}
\end{figure}

Figure \ref{fig:1} shows $\mathtt{Ens}_{\alpha, \tau}$ test performance is more sensitive to $\alpha$ when $\tau$ is large, which makes sense because the ensembler is more like an interpolated sum than a max in this regime. Overall, we find that performance can {\em degrade significantly due to too strong reliance on ID or text} (i.e. $\alpha \in \{0.2, 0.8\}$), but {\em most other settings give near-optimal performance}.

We are also interested in whether careful tuning of $(\alpha, \tau)$ can improve on $\mathtt{EnsRec}$. For Beauty, we {\em do improve} by tuning $(\alpha, \tau)$ on the validation set, leading to test Recall@10 10.00. However, for Steam, the default $(\alpha, \tau) = (0.5, 100)$ {\em already achieves optimal} test Recall@10 (20.45). For both datasets, tuning $(\alpha, \tau)$ on the training data is detrimental since the ID model memorizes more of this data, encouraging too large $\alpha$.

\section{Conclusion}

Our results develop a rigorous understanding of ID-text complementarity in 
dense retrieval-based 
SR. We demonstrate that a simple ID-text complementarity-leveraging method performs surprisingly well, highlighting the potential fruits offered by this complementarity. 
These observations inspire several future directions: (1) studying complementarity along additional axes, e.g. in terms of learning semantic and collaborative signals, and in additional settings, e.g. when training the text encoder \cite{yuan2023go}, 
(2) more sophisticated ensembling to better exploit complementarity, 
e.g. training $(\alpha, \tau)$ or a user-wise gate, (3) training to amplify complementarity, 
and (4) extending $\mathtt{EnsRec}$ to generative recommendation \cite{rajput2023recommender}.

\bibliographystyle{ACM-Reference-Format}
\bibliography{sample-base}

\end{document}